\documentclass[journal, onecolumn, draftcls]{IEEEtran}
\usepackage{macros_sjain}
\usepackage{times}
\usepackage{epsfig}
\usepackage{algorithm}
\usepackage{algorithmic}
\usepackage[normalem]{ulem}

\usepackage{hyperref}

\title{\huge Rank-to-engage: New Listwise Approaches to Maximize Engagement} 
\author{Swayambhoo Jain, Akshay Soni, Nikolay Laptev, and Yashar Mehdad\thanks{This work was done during summer internship at Yahoo! in year 2016.  Swayambhoo Jain is currently a PhD student at University of Minnesota, Twin Cities,  Akshay Soni is a research scientist at Yahoo!, Nikolay Laptev is a senior  scientist at Uber, and Yashar Mehdad is a senior scientist at Airbnb. Author emails: {\tt jainx174@umn.edu,akshaysoni@yahoo-inc.com, nlaptev@uber.com, yashar.mehdad@airbnb.com}. }}

\begin{document}
	\maketitle
\begin{abstract}
For many internet businesses, presenting a given list of items in an order that maximizes a certain metric of interest (e.g., click-through-rate, average engagement time etc.) is crucial. We approach the aforementioned task from a learning-to-rank perspective which reveals a new problem setup.  In traditional learning-to-rank literature, it is implicitly assumed that during the training data generation one has access to the \emph{best or desired} order for the given list of items.  In this work, we consider a problem setup where we do not observe the desired ranking. We present two novel solutions: the first solution is an extension of already existing listwise learning-to-rank technique--Listwise maximum likelihood estimation (ListMLE)--while the second one is a generic machine learning based framework that tackles the problem in its entire generality. We discuss several challenges associated with this generic framework, and  propose a simple \emph{item-payoff} and \emph{positional-gain} model that addresses these challenges. We provide training algorithms, inference procedures, and demonstrate the effectiveness of the two approaches over traditional ListMLE on synthetic as well as on real-life setting of ranking news articles for increased dwell time.
\end{abstract}

\section{Introduction}\label{sec:intro}
Recommending items that matches users interests lies at the core of many online businesses and has been an active area of research; over the years, many techniques have been developed for these tasks including matrix completion based collaborative filtering \cite{candes2009exact,keshavan2009matrix}, factorization machines \cite{rendle2010factorization}, etc. The central theme of these techniques is that they utilize the historical data of user-engagement to predict user's interest or rating for the new items. These items are then presented to the users in the decreasing order of the predicted rating/score. It is implicitly assumed that the decreasing order of predicted score is the best order to show the items to the users. However, our real life experience suggests that in many scenarios user satisfaction is driven not just by the quality of items but also by the order in which they are presented to the users. In scenarios where the intention is better long-term user-engagement or revenue per user, once the most relevant set of items to be shown to the user are identified, the important task is to show these items in an order that maximizes a particular metric of interest, such as average time spent by users per session (each session is a ordered list of items), or total click through rate, etc.

\begin{figure}
	\begin{center}
		\includegraphics[scale=0.33]{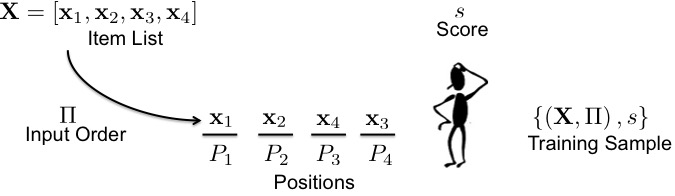}
	\end{center}
	\caption{Our problem setup: A user is served with the list of items $\mathbf{X}$ in the \emph{input} order $\Pi$. The user provides a score $s$ measuring metric of interest for the pair $(\mathbf{X}, \Pi)$.}\label{fig:prob_setup}  
\end{figure}

The problem setup in such scenarios can be abstractly represented as in Figure \ref{fig:prob_setup}, where the list of $n$ items denoted by a feature matrix  $\bX = [\bx_1,\cdots, \bx_n] \in \RR^{d \times n}$ ($\bx_i \in \RR^d$ is the $i^{th}$ column that is the feature vector of the $i^{th}$ item in the list), is shown to a user in the \emph{input} order $\Pi \in \cP_n$ (where $\cP_n$ is the set of all permutations of integers $\{1,\cdots,n\}$). The user assesses the quality of the  items and the input order pair, $(\bX,\Pi)$, and assigns  a score $s \in \RR^+$ that is a measure of desired metric for $(\bX,\Pi)$. This results in a training example $\left\{ \left(\bX,\Pi\right) , s \right\}$. Depending on the specific setting, the user may assign the score explicitly or it may be calculated based on user interaction statistics, for example, in terms of clicks (or no clicks) or the average user engagement times. Using the training examples collected in this manner, the ultimate goal is to predict an order for a new unseen list of items so that the score is maximized. A similar looking problem is the focus point of various techniques developed in learning-to-rank literature but our problem setup is different as it violates an implicit assumption prevalent in learning-to-rank literature: the assumption that \emph{best or desired} order is provided with training data however in our problem setup we do not have such data.

Our main contributions are two machine learning based solutions for the proposed problem setup. The first solution, weighted ListMLE, builds upon a popular listwise learning-to-rank technique ListMLE \cite{xia2008listwise} by incorporating the weights proportional to the scores.  The second solution is a general machine learning framework in which we address the problem at hand in its entire generality. In this we first learn a mapping to predict the score for a given list of items and an order. The final order is obtained by maximizing the predicted score. We reveal several challenges associated with this approach and propose a simple item-payoff and positional-gain model that addresses these specific challenges. We also present an alternating-minimization based training algorithm and demonstrate the effectiveness of the proposed techniques on simulated as well as real datasets. 
\subsection{Related works}
The problem setup we consider lies at the nexus of recommendation and ranking systems. It arises in the context of recommendation systems and is motivated by the learning-to-rank approaches. However, our problem setup is quite different from these traditional settings. As discussed earlier, typically in recommendation systems the items are presented in decreasing order of the predicted ratings. Recommendation techniques that give preference to diversity \cite{ziegler2005improving} or the multi-criteria recommender systems \cite{lakiotaki2011multicriteria} often deviate from ``decreasing order of predicted rating" ranking of items. But even in these systems, the input order is not explicitly modeled as considered in this paper. 

\begin{figure}
\begin{center}
\includegraphics[scale=0.35]{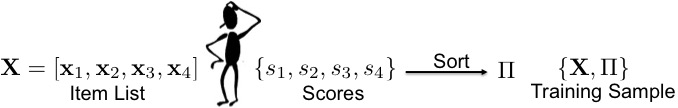}
\end{center}
\caption{\small Existing learning-to-rank setup: The user is shown list of items $\mathbf{X}$ and the user provides the \emph{best or desired} order which may be obtained by decreasing of relevance scores $\{s_i\}_{i=1}^n$.}
  \label{fig:LTR_prob_setup}
\end{figure}

Our problem is also related to the learning-to-rank literature. Traditionally, learning-to-rank problems are motivated from a search engine perspective, where the task of is to show the results in decreasing order of relevance to the user's query. The main goal is to minimize the user's search time. The abstract problem setup arising in learning-to-rank literature is shown in Figure \ref{fig:LTR_prob_setup}, where the user is shown the list of items $\mathbf{X}$ and the user assigns relevance scores $\{s_i\}_{i=1}^n$ to each item for the given query and provides best order based on decreasing order of relevance scores. In this manner training data comprising of lists of items and the final order can be collected. In some cases, however, access to the individual relevance scores is not necessary and the desired order can be inferred using other techniques \cite{li2014learning}.  Variety of machine learning based learn-to-rank algorithms have been developed that use this training data to predict order for a new list of items. The main challenge in applying machine learning to rank list of items is the combinatorial nature of the output domain of the mapping. Existing techniques use different ways to deal with this challenge and can be broadly classified into three main categories: pointwise, pairwise, and listwise ranking \cite{trotman2005learning,liu2009learning,li2014learning}.

The pointwise approaches reduce the problem of ranking to regression tasks. They ignore the combinatorial output domain, and focus on predicting the relevance score of each item separately. Some of the important pointwise techniques are proposed in \cite{cossock2006subset,gey1994inferring,chu2005new,crammer2001pranking} among many others. The pairwise approaches on the other hand reduce the learning-to-rank problem to a classification problem by using pairwise comparisons to transform the order into binary labels. Few notable pairwise approaches among many others include support vector machine (SVM) based approach \cite{Graepel99classificationon}, perceptron based approach \cite{gao2005linear,herbrich1999large} and neural networks based approach \cite{burges2005learning}. Listwise approaches take an entire list of items as input and directly tackle the combinatorial nature of output domain. Due to this, the listwise approaches are known to perform better than the pointwise and pairwise approaches. These are generally based on probabilistic modeling of various orders for the given list of items. Some notable works in listwise learning-to-rank are \cite{xu2007adarank, xia2008listwise, yeh2007learning, volkovs2009boltzrank}. The pursuit to minimize the loss functions defined permutation spaces has lead to several listwise learning-to-rank techniques including LambdaRank \cite{burges2006learning} and several other followup works \cite{from-ranknet-to-lambdarank-to-lambdamart-an-overview}. 

A distinguishing characteristic of traditional learning-to-rank problem setup is that the relevance of items to a query is a property of the items and does not change with the order in which the items are shown to the user. This is the main difference between our problem setup and existing learning-to-rank setup. We have the notion of \emph{input} order whereas no such notion exists in problems discussed in learning-to-rank  literature. Also, in learning-to-rank literature it is implicitly assumed that during the training data generation one has access to the \emph{best or desired} order for the given list of items. The existing learning-to-rank techniques mainly focus on predicting this order in various ways. In our problem setup we do not have access to the \emph{best or desired} order for the items in a list. Due to these reasons  traditional learning-to-rank approaches are incapable of handling our problem setting.  

\subsection{Organization}
Followed by brief discussion on notation in section \ref{sec:notation}, we discuss the problem formulation in section \ref{sec:Prob_formulation}. Section \ref{sec:approach_1} we present our first solution the weighted ListMLE. The second more general approach is proposed in section  \ref{sec:approach_2} and the section \ref{sec:item_payoff_model} describes the item-payoff and 
positional-gain model. The section \ref{sec:experiments} provides experiments to show the efficacy of the proposed approach. Finally, section \ref{sec:conclusions} concludes the paper with a brief discussion on future directions.

\section{Notation}\label{sec:notation}
Vectors and matrices are denoted by bold-face lowercase and uppercase characters, respectively.  A list of size $n$ is represented by the matrix $\mathbf{X} = [\mathbf{x}_1, \cdots, \mathbf{x}_n] \in \RR^{d \times n}$ whose $i^{\rm th}$ column $\mathbf{x}_i \in \RR^d$ is the feature vector of $i^{\rm th}$ item in the list. 
Vectors of all ones and zeros of size $n$ are denoted by $\mathbf{1}_n$ and $\mathbf{0}_n$ respectively. An identity matrix of size $n \times n$ is denoted by $\bI_n$.  The set of all permutations of integers $\{1,\cdots,n\}$ is denoted by $\mathcal{P}_n$. A particular permutation is denoted by $\Pi = [\pi_1, \cdots, \pi_n] \in \mathcal{P}_n$ where $\pi_i$ denotes the position where the $i^{th}$ item  in the list is placed. For example, $\pi_2 = 1$ implies that the second item is placed at the first position. For a given matrix $\bX \in \RR^{d \times n}$ and $\Pi \in \cP_n$,  $\bX_{\Pi}$ denotes a matrix whose columns are obtained by re-ordering columns of $\bX$ as per $\Pi$. The function $\textrm{sort}[x_1,\cdots,x_n]$ returns the permutation denoting the positions of $x_i$'s if they were placed in descending order.

\section{Problem Formulation}\label{sec:Prob_formulation}

As discussed earlier, using the problem setup as shown in Figure \ref{fig:prob_setup}, training data comprising of list of $n$ items $\bX$, the order in which it is shown $\Pi \in \cP_n$, and the user assigned score $s \in \RR^+$ can be collected. Note that there is one single score for the entire list of items. We assume there is a probability distribution $P_{\bX,\Pi,s}$ over $ \RR^{ d \times n } \times \cP_n \times \RR^+$ from which we are given $m$ i.i.d training examples as follows
\begin{align}\label{eq:training_data}
\cD_N = \left\{  \left\{ \left(\bX^{(i)}, \Pi^{(i)} \right), s^{(i)} \right\}_{i=1}^N  \right\}, 
\end{align}
where $\bX^{(i)} \in \RR^{d \times n}$ denotes the $i^{th}$ list of items, $\Pi^{(i)} \in \cP_n$ is the order in which the items were shown to the user, and $s^{(i)}$ is the corresponding score of the list. The goal is to use the training data to learn an ordering for the new list of items such that it maximizes the score. In light of the available training data, addressing this goal is particularly challenging because we do not have access to the order that maximizes score. Next we describe two approaches designed towards to achieve this goal. 

\section{Approach 1: Weighted ListMLE}\label{sec:approach_1}
The main challenge in addressing the problem of ordering a list of items using the training data $\cD_N$ lies in the discrete combinatorial nature of input and output domains. As discussed earlier, the listwise learning-to-rank techniques have effectively addressed this challenge in a related but different setting. Our first approach builds upon a existing popular technique  ListMLE and extends it so that to our problem setting. We first briefly describe the ListMLE technique followed by details of our proposed extension to it. 

\subsection{ListMLE}
The ListMLE approach is based on modeling the conditional probabilities of various permutations given the list of items \cite{xia2008listwise}. Specifically, the conditional probability of a permutation $\Pi \in \mathcal{P}_n$ given the list of items $\mathbf{X} \in \mathbb{R}^{d \times n}$ is modeled by so called Plackett-Luce model as follows \begin{equation}\label{eq:PL}
P(\Pi | \mathbf{X}; g) =  \prod_{j=1}^n \frac{ e^{ g(\mathbf{x}_{\pi_j}) }    }{\sum_{k=j}^n e^{ g(\mathbf{x}_{\pi_k}) } },
\end{equation}
where $g(\cdot): \mathbb{R}^d \rightarrow \mathbb{R}$ computes the score of each item and $\mathbf{x}_{\pi_k}$ denotes the feature vector of  $\pi_k^{th}$ item in the list. Using training data $\cD_N$ the ListMLE entails solving the following maximum likelihood problem 
\begin{equation}\label{eq:ML}
\min_{g} \sum_{i=1}^N - \log \left( P(\Pi^{(i)} | \mathbf{X}^{(i)}; g) \right).
\end{equation}
Note that in ListMLE, it is assumed that the output permutation $\Pi$ is the desired permutation and the goal is to learn a mapping from the feature space to this output space. The learned mapping $\hat g$---the  solution of problem \eqref{eq:ML}---is used to predict the order for a new list of items. For a new list of items $\mathbf{X} \in \mathbb{R}^{d \times n}$, first the predicted relevance scores $\{\hat g(\mathbf{x}_j)\}_{j=1}^n$ are computed. These scores are then used to calculate the probabilities of various permutations using \eqref{eq:PL}. The inference procedure involves finding the maximum probability permutation, which can be efficiently implemented owing to the Plackett-Luce model in \eqref{eq:PL} by sorting the predicted scores $\{\hat g(\mathbf{x}_j)\}_{j=1}^n$. Next, we present our approach which extends ListMLE to our problem setting.

\subsection{Weighted ListMLE}
As discussed earlier, our problem setup has a notion of \emph{input} order. From equation \eqref{eq:ML}, it is clear that ListMLE  allows only one order per list that is assumed to be the \emph{best} order in some sense. The \emph{input} order in our approach may not necessarily be the \emph{best} order as required by ListMLE since we want to figure out that out of all the possible permutations of the items which order corresponds to the best score. We propose weighted ListMLE to address this specific problem setting. Similar to ListMLE, we model the conditional probability of a order $\Pi \in \mathcal{P}_n$ given the input list $\mathbf{X} \in \mathbb{R}^{d \times n}$ by Plackett-Luce model in equation \eqref{eq:PL}. As our aim is to predict the order that maximizes the given metric, we weight the likelihood term for the given list of items and the order in which they were presented by the corresponding score $s$. Specifically, for the training data $\cD_N$, the weighted ListMLE involves solving the following weighted maximum likelihood problem 
\begin{equation}\label{eq:WML}
\min_{g} \sum_{i=1}^N - s^{(i)}\log \left( P(\Pi^{(i)} | \mathbf{X}^{(i)}; g) \right).
\end{equation}
In the above problem, the weights $s^{(i)}$ bias the learning process such that the orders with higher score are given higher probabilities. The algorithm for solving the training problem \eqref{eq:WML} can be shown to be a simple modification to existing training algorithm for ListMLE proposed in \cite{xia2008listwise} by adding weights to the gradient computation.   
After learning the scoring function $\hat g$ by solving the problem in \eqref{eq:WML}, it is used to find the permutation $\Pi$ for a new list $\mathbf{X}$ as follows 
\begin{equation}\label{eq:WML_inference}
\hat \Pi(\mathbf{X}) = \arg \max_{\Pi \in \mathcal{P}_n}  
P(\Pi | \mathbf{X}; \hat g).
\end{equation}
Again, owing to the special structure of Plackett-Luce model, the $\hat \Pi(\mathbf{X})$ can be obtained by simply sorting $  \left\{  e^{\hat g(\mathbf{x}_j) }\right\}_{j=1}^n$ in descending order as follows
\begin{equation}
\hat \Pi(\mathbf{X}) = \textrm{sort} \left[ \exp \left(\hat g(\mathbf{x}_1)\right) ,\cdots, \exp \left(\hat g(\mathbf{x}_n)\right)\right].
\end{equation}
The weighted ListMLE can be reduced to ListMLE if the \emph{input} order $\Pi^{(i)}$ is chosen such that it is based on decreasing order of relevance of the items, i.e. the \emph{best} order, and the corresponding scores is fixed to be constant (say $s^{(i)} = 1$ for all $i$). We extend existing ListMLE in a sense that the notion of \emph{input} order can be accommodated. Weighted ListMLE can be construed as an attempt to extend the existing learning-to-rank to our setting while keeping the essential characteristics of ListMLE intact. Next, we present a more direct approach that handles our problem setup in more generality.

\section{Approach 2: A Machine Learning based Framework}\label{sec:approach_2}
Our ultimate goal it to learn a mapping that maximizes the score for the given list of items. Using machine learning techniques to learn such mapping would require training data in terms of list of items and the score maximizing order and existing learning-to-rank techniques can be applied. But the training data in our problem setup does not have this form that makes developing a machine learning approach to solve this problem challenging. However, using the training data $\cD_N$ in the form it is available to us it is possible to learn a mapping from $\left(\RR^{d \times n}, \cP_n \right)$ to $\RR^+$ because the training data can be considered as noisy observations of such a mapping. Accordingly, we follow a two step approach: (1) learn a mapping that predicts score for the given list of items and order, (2) use the learned mapping to obtain the final order by maximizing the predicted order.  We learn the mapping $f:\left(\RR^{d \times n}, \cP_n \right) \rightarrow  \RR^+$ by solving the following empirical risk minimization problem 
\begin{align}\label{eq:erm}
\hat{f}_{\cD_N} = \arg \min_{ f \in \cF} \ \sum_{i=1}^N  \left(  s^{(i)} - f\left( \bX^{(i)}, \Pi^{(i)} \right)   \right)^2, 
\end{align}
where $\cF$ is the set of functions defined from $\left(\RR^{d \times n}, \cP_n \right)$ to $\RR^+$. For a  new list of items $\bX \in \RR^{d \times n}$ we infer the score maximizing ordering using the learned $\hat{f}_{\cD_N}$ in \eqref{eq:erm} as follows
\begin{align}\label{eq:inference}
\hat \Pi(\bX) = \arg \max_{\Pi \in \cP_n}  \hat{f}_{\cD_N}\left( \bX, \Pi \right).
\end{align}
Above approach draws some parallel from multi-class classification problems where the training data is first used to accurately predict the probability of various classes, and the classifiers output is obtained by maximizing the predicted probability. Here, we first use the training data to fit a function that accurately predict the score for given list of items and input order, and then, use the learned mapping to predict the final order by maximizing the predicted score. The choice of function class $\cF$ is critical to the feasibility of the approach described above as it involves combinatorial input and output domain. Next we discuss various issues that govern the choice of function class $\cF$.

\paragraph{Choosing the function class $\cF$:} As the inference problem in \eqref{eq:inference} involves optimization over set of permutations $\cP_n$, its computational complexity is $ \cO(n!) = \cO(n^n)$ making it computationally prohibitive even for modest values of $n$.  Therefore, the first requirement on the function class $\cF$ is to make the corresponding inference problem in \eqref{eq:inference} feasible. In addition to the inference complexity, note that for the fixed value of list of items $\bX$, the function $f(\bX, \Pi)$ can take $n!$ different values by choosing different $\Pi \in \cP_n$. Therefore, the second requirement on the function class $\cF$ is that it should prevent over-fitting and the estimate $\hat f_{\cD_N}$ should have reasonable variance with practically feasible number of training data points. Both these requirements can be handled if the function class $\cF$ is simple. For these purposes, we propose the class of functions that can be decomposed as follows 
\begin{align}\label{eq:simple_function_class}
f(\bX,\Pi) = \sum_{i=1}^{n} h( \bx_i, \pi_{i}),  
\end{align}
where $h \in \cH$ and $\cH$ is some class of functions defined from $\left( \RR^d, \{1,\cdots,n\} \right)$ to $\RR^+$. The specific structure considered in \eqref{eq:simple_function_class} is simple because the overall score predicted by these functions depends on the item feature vector and the locations it appears in $\Pi$. Further, note that these functions still take $n!$ values for a given $\bX$ by choosing different $\Pi \in \cP_n$. However, each of these values is sum of some $n$ entries chosen from a \emph{scoring} matrix $\bS_h(\bX)$ defined as
\begin{align}\label{eq:scoring_matrix}
\bS_h(\bX) = \begin{bmatrix}
h(\bx_1,1) & \cdots & h(\bx_1,n) \\
\vdots & \cdots & \vdots \\
h(\bx_n,1) & \cdots & h(\bx_n,n) 
\end{bmatrix}.
\end{align}
For a given order $\Pi$, all the terms in the summation in \eqref{eq:simple_function_class} can be obtained from the entries of the scoring matrix $\bS_h(\bX)$. This essentially implies that the functions following the decomposition in \eqref{eq:simple_function_class} have inherent low dimensional structure. 

\paragraph{Training with $\cF$:}
The training with the $\cF$ in \eqref{eq:simple_function_class}, the empirical risk minimization problem in \eqref{eq:erm} reduces to 
\begin{align}\label{eq:training_simple}
\hat{h}_{\cD_N} = \arg \min_{ h \in \cH } \  \sum_{i=1}^N  \left(  s^{(i)} -  \sum_{j=1}^n h(\bx_j^{(i)},\pi_j^{(i)})    \right)^2.
\end{align}
The actual complexity of above training problem will depend on the specific choice of the function class $\cH$. This issue will be discussed in greater details later in this paper when we consider a specific example of $\cH$.

\paragraph{Inference with $\cF$:}
The inference problem in \eqref{eq:inference} reduces to the following problem
\begin{align}\label{eq:inference_simple}
\hat \Pi(\bX) = \arg \max_{\Pi \in \cP_n}  \sum_{i=1}^n \hat{h}_{\cD_N}\left( \bx_i, \pi_i \right).
\end{align}
Further, observing that $\Pi$ is a valid permutation, i.e., at one location only one item is placed, we do a change of variable from permutation $\Pi \in \cP_n$ to a permutation matrix $\bP \in \RR^{n \times n}$. The permutation matrix $\bP$ is such that its entries are either $1$ or $0$ and there is exactly one non-zero entry in each column and row. The rows of $\bP$ can be obtained for a given $\Pi$ in such a manner that if the $i^{th}$ item goes to $j^{th}$ location  $P_{i,j} = 1$. This implies that there is one to one mapping from $\Pi$ to $\bP$ and the objective in problem \eqref{eq:inference_simple} can be written in terms of $\bP$ as follows 
\begin{align*}
\sum_{i=1}^n \hat{h}_{\cD_N}\left( \bx_i, \pi_i \right) = \sum_{i=1}^n \sum_{j=1}^n P_{ij} \hat{h}_{\cD_N}\left( \bx_i, j \right).
\end{align*}
Next we use the notion of scoring matrix introduced in \eqref{eq:scoring_matrix} and we introduce analogous scoring matrix $\bS_{\hat{h}_{\cD_N}}(\bX)$ whose $(i,j)^{th}$ entry is $\hat{h}_{\cD_N}\left( \bx_i, j \right)$. With this the problem the inference problem in \eqref{eq:inference_simple} can be converted to an equivalent problem as follows
\begin{equation}\label{eq:integer_program}
\begin{aligned}
& \underset{\bP \in \RR^{n \times n}}{\text{min}} 
& & \textrm{Tr}(  \mathbf{P} \bS_{\hat{h}_{\cD_N}}(\bX)  ) \\
&\text{subject to} && \ \sum_{i=1}^n P_{ij} = 1, \ \forall j, \ \sum_{j=1}^n P_{ij} = 1  \ \forall i,\\
& & &   P_{ij} \in \{0,1\} \ \forall i,j,
\end{aligned}
\end{equation}
where $\textrm{Tr}( \cdot )$ represents sum of diagonal entries of a matrix. Problem \eqref{eq:integer_program} is an instance of the classical \emph{linear sum assignment problem} that due to the \emph{total unimodularity} of the constraints can be efficiently solved by relaxing it to the following linear program \cite{burkard1980linear}
\begin{equation}\label{eq:linear_program}
\begin{aligned}
& \underset{\bP \in \RR^{n \times n}}{\text{min}} 
& & \textrm{Tr}(  \mathbf{P} \bS_{\hat{h}_{\cD_N}}(\bX)  ) \\
&\text{subject to} && \ \sum_{i=1}^n P_{ij} = 1, \ \forall j, \ \sum_{j=1}^n P_{ij} = 1  \ \forall i,\\
& & &   P_{ij} \ge 0 \ \forall i,j.
\end{aligned}
\end{equation}
Above inference problem is a linear program of $n^2$ variables that can be solved in polynomial time as compared to the original inference problem in \eqref{eq:inference} whose complexity without our choice of simpler function class $\cF$ could be $\cO(n^n)$ in worst case. Recently, a fast greedy algorithm with provable $\frac{1}{2}-$optimal solution with a worst-case runtime of $\cO(n^2)$ was used in \cite{shah2017} for online constrained ranking problems. 


\section{An instance of $\cF$: The item-payoff and positional-gain model}\label{sec:item_payoff_model}
Here we propose a specific instance of the class $\cF$ that follows the decomposition in \eqref{eq:simple_function_class}. The proposed model utilizes the notions of positional-gains and item-payoffs. For the given list of items $\bX$, the item-payoff vector whose $i^{th}$ entry denotes the payoff associated with $i^{th}$ item is modeled as follows
\begin{align}\label{eq:item_payoff_vector}
\exp\left( \bX^T\bv^* \right) =  \begin{bmatrix} \exp \left( \mathbf{x}_1^T  \mathbf{v}^* \right), \cdots ,\exp \left( \mathbf{x}_n^T  \mathbf{v}^* \right)   \end{bmatrix}^T 
\end{align}
where $\mathbf{v}^* \in \RR^d$ is a fixed ground truth weight vector. 
The positional-gain is the property of position and it is defined by the gain vector $\mathbf{g}^* \in \mathbb{R}^n$ whose  $i^{th}$ component $g_i$ denotes the gain associated with $i^{th}$ position. With this for the given list of items $\mathbf{X}$ and order $\Pi \in \mathcal{P}_n$, the score is calculated as follows 
\begin{equation}\label{eq:quality_metric_1}
f(\Pi,\mathbf{X}) = \left(\bg^*\right)^T \exp\left( \bX^T_{\Pi}\bv^* \right),  
\end{equation}
where $\bX_{\Pi}$ is the matrix whose columns are obtained by ordering columns of $\bX$ as per $\Pi$. The function in \eqref{eq:quality_metric_1} is an instance of the function class defined in \eqref{eq:simple_function_class} with $h(\bx_i,\pi_i) = g_{ \pi_i}^*\exp\left(\bx_i^T\bv^*\right)$. A similar model was proposed in \cite{vanchinathan2014explore} in context of explore and exploit in top-N recommender systems however it focused on modeling the item relevance under the assumption that first position is more important than second and so on. In contrast, we do not have such an assumption in our problem setup.

\paragraph{Training:} As $h(\bx_i,\pi_i) =  g_{\pi_i}^*\exp\left(\bx_i^T\bv^*\right)$, the function class $\cH$ is parametrized by the positional-gain vector $\bg$ and the weight vector $\bv$. With this the empirical risk minimization problem in \eqref{eq:training_simple} reduces to
\begin{align}\label{eq:erm_payoff_gain}
\min_{\bv \in \RR^d, \bg \in \RR^n} \sum_{i=1}^N \left( s^{(i)} -   \exp \left( \bv^T \bX^{(i)}_{\Pi^{(i)}} \right)  \bg  \right)^2.
\end{align}
The above problem suffers from scaling ambiguity due the product term $\exp \left( \bv^T \bX^{(i)}_{\Pi^{(i)}} \right)  \bg $. In addition, this term increases exponentially with scaling of $\bv$ which results in numerical overflow issues. For these purposes instead of solving problem \eqref{eq:erm_payoff_gain} we solve the following modified problem for training 
\begin{equation}\label{eq:erm_payoff_gain_modified}
\begin{aligned}
& \underset{\bv \in \RR^d, \bg \in \RR^n}{\text{min}} 
 \sum_{i=1}^N \left( s^{(i)} -   \exp \left( \bv^T \bX^{(i)}_{\Pi^{(i)}} \right)  \bg  \right)^2 +  \lambda \|\bg\|_2^2 \\
&\text{subject to} \quad  \| \bv \|_2 \le 1.
\end{aligned}
\end{equation}
where $\lambda>0$ is a regularization parameter. Even though we have addressed the issue of scaling ambiguity the problem in \eqref{eq:erm_payoff_gain_modified} is still jointly non-convex in $\bg$ and $\bv$. However, for a fixed $\bv$ the problem is convex in 
$\bg$ and similarly, for a fixed $\bg$ as well the problem is convex in $\bv$. Based on this we propose an alternating minimization based algorithm for approximately solving problem \eqref{eq:erm_payoff_gain_modified}. Starting with initial $\bv^{(0)} = \frac{\mathbf{1}_n}{\sqrt{n}}$ we alternatively minimize with respect to $\bg$ and $\bv$ until convergence. The final procedure is detailed in Algorithm \ref{ag:alt_min}.
\begin{algorithm}[!t]
	\caption{Alternating minimization training algorithm for item-payoff and positional-gain model.}
	\begin{algorithmic} 
		\STATE \hspace{-1.2em} \textbf{Inputs:}  Training Data: $\{ \left( \mathbf{X}^{(i)}, \Pi^{(i)} \right),  s^{(i)} \}_{i=1}^N$  and  $\epsilon$.
        \STATE \hspace{-1.2em} Initialize: $\bv^{(1)} = \mathbf{1}_n/\sqrt{n}$, $\text{obj}^{(1)} = \sum_{i=1}^N \left(s^{(i)}\right)^2$.
        \STATE \hspace{-1.2em} \textbf{Repeat:} $k=1,\cdots $ 
        \STATE  Update $\bg^{(k+1)}$ by solving: 
        \begin{align*}
\min_{\mathbf{g} \in \RR^n}   \sum_{i=1}^N &\left( s^{(i)} - \exp \left(  \left(\bv^{(k)} \right)^T \bX^{(i)}_{\Pi^{(i)}} \right)  \bg \right)^2 + \lambda \| \mathbf{g} \|_2^2
		\end{align*}
        
        \STATE Update $\bv^{(k+1)}$ by solving  using Algorithm \ref{ag:gradient_proj}:
       \begin{equation*}
\begin{aligned}
& \underset{\bv \in \RR^d}{\text{min}}
&& \sum_{i=1}^N \left( s^{(i)} - \exp \left(  \bv^T \bX^{(i)}_{\Pi^{(i)}} \right)  \bg^{(k+1)} \right)^2  \\
& \text{subject to} &&   \ \| \bv \|_2 \le 1.
\end{aligned}
\end{equation*}
\STATE Calculate: 
\begin{align*}
\text{obj}^{(k+1)} = \sum_{i=1}^m & \left( s^{(i)} - \exp \left(  {\bv^{(k+1)}}^T \bX^{(i)}_{\Pi^{(i)}} \right)  \bg^{(k+1)} \right)^2   \\ & + \lambda \|\bv^{(k+1)}\|_2^2
\end{align*}
\STATE \hspace{-1.4em} \textbf{Until:} $ \text{obj}^{(k)}- \text{obj}^{(k+1)} \le \epsilon $. 
\STATE \hspace{-1.4em} \textbf{Output:} $\hat{\mathbf{v}} = \mathbf{v}^{(k)}, \hat{\mathbf{g}} = \mathbf{g}^{(k)}$. 
	\end{algorithmic} \label{ag:alt_min}
\end{algorithm}
The $\bg$ update step in this algorithm is a standard $\ell_2$ regularized least squares problem which can solved in closed form and the $\bv$ update step is a constrained convex program which can be solved by a projected gradient descent approach shown in Algorithm \ref{ag:gradient_proj}. 

\begin{algorithm}[t!]
	\caption{Projected gradient descent for $\bv$ update.}
	\begin{algorithmic} 
		\STATE \hspace{-1.4em} \textbf{Inputs:} $\{ \left( \mathbf{X}^{(i)}, \Pi^{(i)} \right),  s^{(i)} \}_{i=1}^N$, $\bg$, $\eta$,  $\epsilon$.
        \STATE \hspace{-1.4em} Initialize: $\mathbf{v}^{(1)} = \mathbf{0}_n$.
        \STATE \hspace{-1.4em} \textbf{Repeat:} $k=1,\cdots $ 
		\STATE  Calculate gradient: 
       \begin{align*}
       	\mathbf{d}^{(k)}= \sum_{i=1}^N 2 e_i^{(k)} \bX^{(i)}_{\Pi^{(i)}} \textrm{Diag}(\bg)  \exp \left( \left(\bX^{(i)}_{\Pi^{(i)}}\right)^T {\bv^{(k)}}  \right)  
		\end{align*}        
        where  $e_i^{(k)} =  \exp \left(  {\bv^{(k)}}^T \bX^{(i)}_{\Pi^{(i)}} \right)  \bg    - s^{(i)}$
\STATE  Update: $\mathbf{v}^{(k+1)}  =  \mathbf{u}^{(k)} - \eta \mathbf{d}^{(k)}$.  
\STATE Project: $\mathbf{v}^{(k+1)} = \min \left\{ 1 , \frac{ 1}{\| \mathbf{v}^{(k+1) }  \|_2} \right\}  \mathbf{v}^{(k+1)}$ 
        \STATE \hspace{-1.4em} \textbf{Until:} $ \| \bv^{(k+1)} - \bv^{(k)}   \|_2 \le \epsilon $.
        \STATE \hspace{-1.4em} \textbf{Output:} $\mathbf{v}^{(k+1)}$.
	\end{algorithmic} \label{ag:gradient_proj}
\end{algorithm}

\paragraph{Inference:} After obtaining $\hat \bg, \hat \bv$ from Algorithm \ref{ag:alt_min}, they can be used to obtain an estimate for the score for the given list of items $\bX$ and input order $\Pi$ as follows
\begin{align}\label{eq:predicted_score}
\hat f(\Pi,\bX) = \left( \hat \bg \right)^T \exp\left(  \bX_\Pi^T \hat \bv  \right).
\end{align}
For inferring the order that maximizes the predicted score we first use the fact that for the item-payoff and positional-gain model, $h(\bx_i,\pi_i) = \hat{g}_{\pi_i} \exp\left( \bx_i^T \hat{\bv} \right)$ and calculate the scoring matrix followed by solving the linear program in \eqref{eq:linear_program}. However, owing to the linear structure of the positional-gain and item-payoff model the score maximizing order simply corresponds to first sorting the estimated payoffs $\left\{ \exp\left( \bx_i^T \hat{\bv} \right) \right\}_{i=1}^n$ for each item and then putting the item with largest estimated payoff at the position with largest estimated gain and so on.

\section{Experiments}\label{sec:experiments}
We evaluate our approach on the synthetic as well as real data.
\subsection{Synthetic Data}
\begin{table*}
\centering
\begin{tabular}{|c|c|c|c|}
\hline
\textbf{Gain vector $\mathbf{g}$} & \textbf{ListMLE} &   \textbf{Weighted ListMLE} & \textbf{Item-payoff Positional-gain  model}  \\ 
\hline
$[0.2,  0.2,  0.2,  0.2,  0.2]$ & $4.2488$ & $4.2488$  &  $4.2488$  \\
\hline
$[ 0.00493, 0.00493,  0.493,  0.493,  0.00493]$ & $4.0802$  &  $4.7507$ & $5.6239$  \\
\hline
$[0.1667,  0.04167,  0.25, 0.4167, 0.1250]$ & $3.5360$  & $3.8721$  & $4.4252$ \\
\hline
\end{tabular}
\caption{Synthetic data experiment results for different gain vectors.} 
\label{tab:synthetic results} 
\end{table*}

For the synthetic experiments, we fixed the list size as $n=5$ and dimensionality of feature vector as $d=10$. The mean vectors $\{ \mathbf{\mu}_i \}_{i=1}^{n}$ for each item were generated once at start of the experiment such that their components are i.i.d. random variable uniformly distributed in the interval $[0,1]$. A random list of items is generated such that the feature vector for $i^{th}$ item in the list follows a multivariate Gaussian distribution $\mathcal{N} \left( \mathbf{\mu}_i,  \mathbf{I}_n/10 \right)$. Further, using a randomly generated vector $\bv^* \in \RR^n$ generated once at start of the experiment the score for given input order $\Pi$ and list of items $\bX$
was generated as follows
\begin{align}\label{eq:quality_metric}
s(\Pi,\bX) = \left(\bg^*\right)^T \left[ \frac{\exp\left( \bX^T_{\Pi}\bv^* \right)}{\mathbf{1}_n^T\exp\left( \bX^T_{\Pi}\bv^* \right) } \right], 
\end{align}
where $\bg^* \in \RR^n$ is a fixed positional-gain vector. This serves as a ground truth model for calculating the score. Note this score calculation does not exactly follow the item-payoff positional gain model.
For training $N=1000$ lists were randomly generated and input order for each list was chosen uniformly at random from the set $\mathcal{P}_n$ and corresponding score was calculated using \eqref{eq:quality_metric} to obtain the training data $\left\{ \left(\left( \mathbf{X}^{(i)}, \Pi^{(i)} \right),  s^{(i)}\right) \right\}_{i=1}^N$. 
For training weighted ListMLE we fixed the function $g(\bx) = \bu^T \bx$ where $\bu \in \RR^d$,  and solved \eqref{eq:WML} to obtain $\hat \bu$. With linear $g(\bx)$ the problem in \eqref{eq:WML} can be shown to be a convex program that can be efficiently solved by gradient descent algorithm. For a new list the final order was obtained using \eqref{eq:WML_inference} with  $\hat g(\bx)  = \hat \bu^T \bx$. For the second approach we used training data along with Algorithm \ref{ag:alt_min} to obtain $\hat \bv$ and $\hat \bg$. For a new list these vectors were used to obtain the final order by maximizing the predicted score in equation \eqref{eq:predicted_score}.

We compare our approaches to  ListMLE that requires access to desired order, i.e., the order that maximizes the score in \eqref{eq:quality_metric}. However, this is not possible in above experimental setting. Typical ListMLE would use an order obtained by sorting the per-item relevance scores. Here, we provided the relevance score vector for items in the list as $\mathbf{y}^{(i)} = \left( \mathbf{X}^{(i)} \right)^T \mathbf{v^*}$ whose were components sorted to obtain the training data as $\left\{ \left( \mathbf{X}^{(i)}, \Pi^{(i)}_2 = \textrm{sort}(\mathbf{y}^{(i)})  \right)\right\}$. Here too, we fixed $g(\bx) = \bw^T \bx$ and solved (3) to obtain $\hat \bw$. For a new list of items $\hat{\bw}$ was used to predict the final order by  permutation by maximizing the probability in \eqref{eq:PL}.
	 
For testing, $500$ new lists of items were randomly generated and for each list in the test data, orders predicted by all the approaches were obtained and respective scores were calculated using the ground truth model in \eqref{eq:quality_metric}. We repeated the experiments for three different positional-gain vectors and the average scores with various approaches are shown in Table \ref{tab:synthetic results}. The $1^{\rm st}$ row represents the case when all the positions have same gain, i.e., this implies there is no positional preference and as expected we see that all approaches perform the same. The $2^{\rm nd}$ row represent the case with positional-gain vector is skewed so that only third and fourth position are important whereas the $3^{\rm rd}$ row represents a less skewed position gains.  We observe that our approached performs better than ListMLE and the second approach performs the best in both these cases. The superior performance of our approaches can be attributed to the fact that they model input order explicitly. 

The main reason for experiments on synthetic data was to understand the effectiveness of proposed solutions in an ideal setting where input order effect can be precisely controlled. Using various positional gain vectors we were able to show that empirically our proposed approaches are successful as compared to the traditional ListMLE.
We would also like to acknowledge that there may be many other ways of  generating the score $s(\mathbf{X}, \Pi)$ but for the purposes of demonstrating the main idea we choose the specific model in \eqref{eq:quality_metric}. Our main goal here is to highlight a setting where the desired order is not available during training and sorting according to the relevance scores may not be the best thing. We note that our main critique is not that of a particular learning-to-rank technique but its problem setting and ListMLE just serves as an popular representative example of that problem setting.  Next, we demonstrate the effectiveness of our approaches in a real-life setting.

\subsection{Real data}
\begin{table*}[h]
	\centering
	\begin{tabular}{|c|c|c|c|}
		\hline
		& \textbf{ListMLE} & \textbf{Weighted ListMLE} &\textbf{Item-payoff Positional-gain  model }  \\ 
		\hline
		
		\textbf{Avg. NDCG}   &  $ 0.7876$ & $0.8138 $ & $ 0.8516$ \\
		\hline
			\textbf{Top-1 Avg. Dwell time in seconds}  & $ 367.463$  &  $ 405.373$ & $ 464.23$ \\
			\hline
	\end{tabular}
	\caption{Real-world data experiment results.} 
	\label{tab:real_data_results} 
\end{table*}


For the experiments with real data we used data from Yahoo! (www.yahoo.com) that is predominantly a news website; the items in this setup are the news articles. Each article can be related to a few content-categories out of a total of $405$ categories internally defined by Yahoo!. For instance, an article can have a score of $0.5$ towards the category politics along with a score of $0.1$ towards entertainment. Association of articles to these categories is part of Yahoo's content understanding platform whose details are out of the scope of this paper. But as a outcome of this content ingestion and understanding pipeline, each article is represented by a feature vector in $d = 405$ dimensional space. Each user is served a list of articles and the order in which these articles are presented is captured by our training data. The size of list was fixed as $n=3$. The data was collected using logs obtained over one day of website usage. From the resulting logs, we obtained the list of news articles and their feature representations, the order in which they were presented, and the corresponding dwell time or average time the user spent on the entire list. The metric of interest here is the dwell time.  After preprocessing the data we obtained a total of $4950$ data points out of which we used $4000$ examples for training and rest for testing. 

We note that the dwell time is affected by the relative position of news articles with respect to each other. In this particular real life application there is no clear notion of per item relevance rather we just have the dwell time which is a function of the list of news articles and the order in which they were shown on the website. This is an example where traditional learning-to-rank approaches are not applicable. We, however, apply the ListMLE where the training data for ListMLE was fixed as the list of news articles and order in which they were shown using the existing ranking mechanism. In this manner ListMLE learns to predict order as per the current ranking system on the news website.    

We note here that accessing the quality of ordering given by various approaches is a bit tricky because in the test set the dwell times corresponding to all the permutations of given list of items is not available. In other words, we only have partial ground truth order available to us. In order to deal with this, we first found the various orders that available for a the given list of news article in the test data and noted their dwell times. We then ranked these orders for given list of items based on their dwell times. For the same list of news articles we calculated the predicted scores for these orders using item-payoff and positional gain model, and probabilities in case of weighted ListMLE and ListMLE.   Finally, the Normalized discounted cumulative gain (NDCG) score between the ranking of orders obtained by decreasing order of dwell time from test data and the ranking of orders obtained by different approaches was calculated. The goal here is to check whether our approaches gives higher score to the orders with higher dwell time.  We also calculated the average dwell for the top order (among the orders available in the test data) predicted by all the approaches.
The final results are shown in Table \ref{tab:real_data_results}. The reported results averaged over $10$ random splitting of data in training and test sets. We observe that our approach performs better than ListMLE in terms of average NDCG and average dwell times for the top-1 order predicted by our approaches.  The item-payoff and positional-gain model based approach performs the best. These results show that our approach predicts order that correlates more with the order as per the dwell times.  
The relative performances of these three approaches can be understood based on how they model the \emph{input}-order. ListMLE that does not model the input order performs the worst followed by weighted ListMLE that can construed an minor modification to ListMLE whereas the item-payoff and positional-gain approach explicitly models the \emph{input}-order and performs the best.



\section{Conclusion and Future Directions}\label{sec:conclusions} 

In this paper, we investigate the problem of ranking list of items to maximize a given metric of interest when the \emph{best or desired} order is not provided during the training. Following the learning-to-rank based 
route to solve this problem we reveal a new problem setup that is usually not considered in traditional learning-to-rank literature. We proposed two approaches: (1) weighted ListMLE and (2) a generic machine learning framework and item-payoff and positional-gain as an instance of the generic framework. The effectiveness of the proposed approaches was demonstrated on simulated as well as real-life setting of ranking news articles for increased dwell time. 

Future directions for this work include establishing the sample complexity bounds and generalization guarantees for the proposed approaches. Exploring more complex models than item-payoff and positional-gain model is yet another interesting direction for future research. 


 
\bibliographystyle{IEEEbib}
\bibliography{references}

\end{document}